\documentclass[conference]{IEEEtran}
\IEEEoverridecommandlockouts
\usepackage{times,float,multirow,amssymb,amsthm,amsmath,url,graphicx,booktabs,subfigure,graphicx,cite}
\usepackage[lined,ruled,commentsnumbered]{algorithm2e}
\newtheorem{dfn}{Definition}
\def\BibTeX{{\rm B\kern-.05em{\sc i\kern-.025em b}\kern-.08em
    T\kern-.1667em\lower.7ex\hbox{E}\kern-.125emX}}

\begin{document}

\title{Discriminative Joint Probability Maximum Mean Discrepancy (DJP-MMD) for Domain Adaptation\\
\thanks{This research was supported by the Hubei Technology Innovation Platform under Grant 2019AEA171 and the National Natural Science Foundation of China under Grant 61873321.} }

\author{\IEEEauthorblockN{Wen Zhang and Dongrui Wu}
\IEEEauthorblockA{School of Artificial Intelligence and Automation, Huazhong University of Science and Technology, Wuhan, China
\\ Email: \{wenz,drwu\}@hust.edu.cn}
}

\maketitle

\begin{abstract}
Maximum mean discrepancy (MMD) has been widely adopted in domain adaptation to measure the discrepancy between the source and target domain distributions. Many existing domain adaptation approaches are based on the joint MMD, which is computed as the (weighted) sum of the marginal distribution discrepancy and the conditional distribution discrepancy; however, a more natural metric may be their joint probability distribution discrepancy. Additionally, most metrics only aim to increase the transferability between domains, but ignores the discriminability between different classes, which may result in insufficient classification performance. To address these issues, discriminative joint probability MMD (DJP-MMD) is proposed in this paper to replace the frequently-used joint MMD in domain adaptation. It has two desirable properties: 1) it provides a new theoretical basis for computing the distribution discrepancy, which is simpler and more accurate; 2) it increases the transferability and discriminability simultaneously. We validate its performance by embedding it into a joint probability domain adaptation framework. Experiments on six image classification datasets demonstrated that the proposed DJP-MMD can outperform traditional MMDs.
\end{abstract}


\section{Introduction}

A basic assumption in statistical machine learning is that the training and the test data are from the same distribution. However, this assumption does not hold in many real-world applications. Additionally, annotating data for a new domain is often expensive and/or time-consuming; thus, there often exists a challenge that we have plenty of data, with very limited or even no labels \cite{pan2009survey}.

Domain adaptation (DA), or transfer learning, has shown promising performance in handling these challenges  \cite{pan2011domain,gong2012geodesic,hou2016unsupervised,long2017deep,zhang2017joint,lu2018embarrassingly,drwuEA2020}, by transferring knowledge from a labeled source domain to a new unlabeled or partially labeled target domain. It has been widely used in image classification \cite{gopalan2011domain,long2013transfer}, emotion recognition \cite{Ng2015Deep}, brain-computer interfaces \cite{drwuTHMS2017,drwuTNSRE2016}, and so on.

According to \cite{pan2009survey}, DA can be applied when the source and the target domains have different feature spaces, label spaces, marginal probability distributions, and/or conditional probability distributions. Conventional DA approaches follow this assumption, and they mainly use some metrics to separately measure the marginal and/or conditional probability distribution discrepancies. However, the distribution discrepancy of two domains may be better measured by the joint probability distributions. This paper considers directly the case that the source and the target domains have different joint probability distributions, and proposes an approach to compute the corresponding discrepancy.

The most popular DA is feature-based \cite{pan2009survey,long2013transfer,zhang2017joint}, which projects different domains' data into a shared subspace to minimize their discrepancy, usually measured by maximum mean discrepancy (MMD) \cite{gretton2012kernel}. DA may minimize the marginal MMD only \cite{pan2011domain}, or both the marginal and the conditional MMDs with equal weight \cite{ding2018graph} or different weights \cite{wang2018visual}, and has been used in statistical machine learning, deep learning \cite{ghifary2014domain,yan2017mind}, and adversarial learning \cite{ganin2016domain}.

Joint distribution adaptation (JDA) \cite{long2013transfer} is a popular DA approach, which measures the distribution shift between domains by a joint MMD, which includes both the marginal and the conditional MMDs. For joint MMD based approaches, the marginal and conditional distributions are often treated equally, which may not be optimal. So, balanced DA and dynamic DA (both are called BDA in this paper) were proposed to give them different weights by grid search \cite{wang2017balanced} or $\mathcal{A}$-distance \cite{wang2018visual}. However, both the joint and the balanced MMDs compute the discrepancy between two domains as the sum of the marginal and the conditional distribution discrepancies, whereas the joint probability distribution discrepancy may be a better choice, from a Bayesian Theorem perspective.

Additionally, to facilitate DA, two measures need to be considered during feature transformation \cite{chen2019transfer}. The first is \emph{transferability}, which minimizes the discrepancy of the same class between different domains. The other is \emph{discriminability}, which maximizes the discrepancy between different classes of different domains, and hence different classes can be more easily distinguished. Traditional distribution adaptation approaches \cite{long2013transfer,cao2018unsupervised} consider the transferability only but ignore the discriminability.

\begin{figure}[htpb]  \centering
\includegraphics[width=0.50\textwidth,trim=15 4 6 5,clip]{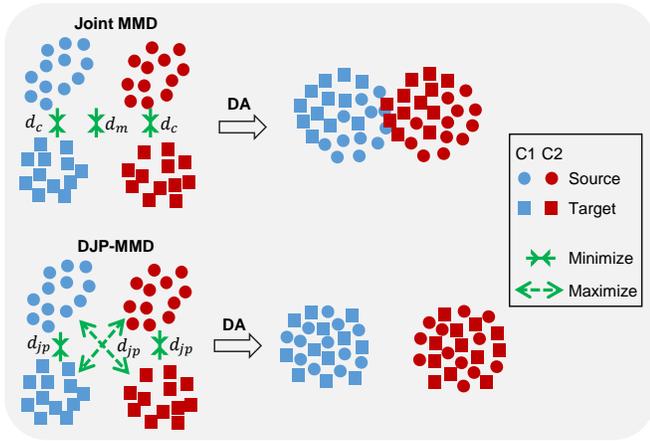}
\caption{Comparison between the traditional joint MMD and the proposed DJP-MMD in DA. The solid lines mean minimizing the marginal ($d_m$), conditional ($d_c$), or joint probability ($d_{jp}$) discrepancies for improved transferability. The dash lines mean maximizing the joint probability discrepancies ($d_{jp}$) between different classes for improved discriminability. When used in DA, DJP-MMD makes the same class from different domains more consistent, and different classes more separated, which facilitate classification.} \label{fig:overview}
\end{figure}

In this paper, we propose discriminative joint probability MMD (DJP-MMD) for DA, which simultaneously minimizes the joint probability distribution discrepancy of the same class between different domains for transferability, and maximizes the joint probability distribution discrepancy between different classes of different domains for discriminability. DJP-MMD can also be easily kernelized to consider nonlinear shifts between different domains. Fig.~\ref{fig:overview} illustrates the difference between the traditional MMD and DJP-MMD.

We validated the performance of DJP-MMD by embedding it into a joint probability domain adaptation (JPDA) framework with simple regularization. Extensive experiments on six real-world image classification datasets demonstrated its superior performance over traditional MMDs.

In summary, our main contributions are:
\begin{itemize}
\item We provide a new theoretical basis for computing the discrepancy between two domains, by considering the joint probability distribution discrepancy directly, which is more accurate and easier to compute.
\item We propose a novel DJP-MMD, which simultaneously maximizes the between-domain transferability and the between-class discriminability for better DA performance.
\item We conduct extensive experiments to demonstrate the advantage of the proposed DJP-MMD over traditional MMDs.
\end{itemize}

\section{Related Work}

Our work is mainly related to traditional MMD based DA, e.g., JDA and BDA. This section briefly reviews them.

\subsection{Joint Distribution Adaptation (JDA)}

Long \emph{et al.} \cite{long2013transfer} proposed joint MMD to measure the discrepancy between two domains in a reproducing kernel Hilbert space (RKHS), using both the marginal and the conditional MMDs:
\begin{align}\label{eq:JDA}
\begin{split}
d(\mathcal{D}_s,\mathcal{D}_t)\approx  &d(P(X_s),P(X_t))\\
&+d(P(Y_s|X_s),P(Y_t|X_t)),
\end{split}
\end{align}
where $\mathcal{D}_s$ and $\mathcal{D}_t$ denote the source and the target domain distribution, respectively, and $d$ is an MMD metric. JDA ignores the relationship between different conditional distributions, and also the dependency between the marginal and the conditional distributions.

\subsection{Balanced Distribution Adaptation (BDA)}

The balanced MMD, originally introduced in \cite{wang2017balanced}, uses grid search to find the weights of the marginal and conditional MMDs. However, this cannot be performed in DA applications that do not have validation sets. Wang \emph{et al.} \cite{wang2018visual} then proposed to use the $\mathcal{A}$-distance \cite{ben2007analysis} to estimate the weights.

This paper considers only the $\mathcal{A}$-distance based BDA, which matches the marginal and the conditional distribution between two domains with a trade-off parameter $\mu \in [0,1]$:
\begin{align}\label{eq:BDA}
\begin{split}
d(\mathcal{D}_s,\mathcal{D}_t)\approx &(1-\mu)d(P(X_s),P(X_t))\\
&+\mu \cdot d(P(Y_s|X_s),P(Y_t|X_t)).
\end{split}
\end{align}

For $C$-class classification, the weight $\mu$ is estimated by:
\begin{align}
\mu \approx 1-\frac {d_m} {d_m+\sum_{c=1}^Cd_c},
\end{align}
where $d_m$ (or $d_c$) equals $2(1-2\epsilon(f))$, in which $f$ is the error of training a linear classier $f$ discriminating all samples from the two domains $\mathcal{D}_s$ and $\mathcal{D}_t$ (or samples in Class~$c$ of the two domains).

Unfortunately, as shown later in our experiments, BDA cannot guarantee performance improvements over JDA. Additionally, BDA needs to train $C+1$ classifiers to calculate $\mu$, which may be computationally expensive for big data.

\section{The Proposed DJP-MMD}

Given a source domain $\mathcal{D}_s$ with $n_s$ labeled samples $\left\{X_s, Y_s\right\}=\{\left(\mathbf{x}_{s, i}, y_{s, i}\right)\}_{i=1}^{n_s}$, and a target domain $\mathcal{D}_t$ with $n_t$ unlabeled samples $X_t=\{\mathbf{x}_{t, j}\}_{j=1}^{n_t}$, where $\mathbf{x}\in \mathbb{R}^{d\times1}$ is the feature vector, and $y$ is its label, with $y\in \{1,\cdots,C\}$ for $C$-class classification. Assume the feature spaces and label spaces of the two domains are the same, i.e., $\mathcal{X}_s=\mathcal{X}_t$ and $\mathcal{Y}_s=\mathcal{Y}_t$, which is a common assumption in homogeneous transfer learning. DA seeks to learn a mapping $h$ that brings $h(X_s)$ and $h(X_t)$ together, so that a classifier trained on $h(X_s)$ can also work well on $h(X_t)$. Different from previous DA approaches, we do not assume $P(X_s) \neq P(X_t)$ or $P\left(Y_s | X_s\right) \neq P\left(Y_t | X_t\right)$ separately; instead, we assume $P(X_s,Y_s) \neq P(X_t,Y_t)$ directly.

Consider a mapping $h$ that maps $\mathbf{x}$ to a lower-dimensional subspace. The general objective function of DA is:
\begin{align}
\min_{h} \ d_{S,T}+\lambda \mathcal{R}(h), \label{eq:obj}
\end{align}
where $d_{S,T}=d(P(X_s,Y_s),P(X_t,Y_t))$ is a discrepancy metric between the source and target domain distributions, $\mathcal{R}(h)=\|h\|_F^2$ controls the mapping complexity, and $\lambda$ is a regularization parameter.

\subsection{Revisit the Traditional MMD Metric}

In traditional feature-based DA, MMD is frequently adopted to measure the distribution discrepancy between the source and the target domains.

A distribution is completely described by its joint probability $P(X,Y)$, which can be equivalently computed by $P(Y|X)P(X)$ or $P(X|Y)P(Y)$. The traditional MMD, e.g., (\ref{eq:JDA}) and (\ref{eq:BDA}), can be summarized as
\begin{align}
\begin{split}
d(\mathcal{D}_s,\mathcal{D}_t)&=d(P(Y_s|X_s)P(X_s),P(Y_t|X_t)P(X_t))\\
&\approx \mu_1 d(P(X_s),P(X_t))\\
& + \mu_2 d(P(X_s|Y_s),P(X_t|Y_t)),   \label{eq:JDA0}
\end{split}
\end{align}
which is a two-step approximation of the joint probability distribution discrepancy \cite{long2013transfer}. First, it uses $P(Y|X)+P(X)$ to estimate $P(Y|X)P(X)$. This ignores the dependency between $P(Y|X)$ and $P(X)$. Second, it uses the class-conditional distribution $P(X|Y)$ to estimate the posterior probability distribution $P(Y|X)$, since the latter is difficult to compute.

Let $h$ be the feature mapping function of $\mathbf{x}$. Then, we adopt the projected MMD \cite{quanz2009large} and compute the marginal distribution discrepancy as $d(P(X_s),P(X_t))=\|\mathbb{E}[h(\mathbf{x}_s)]-\mathbb{E}[h(\mathbf{x}_t)]\|^2$, and the conditional distribution discrepancy as $d(P(X_s|Y_s),P(X_t|Y_t))=\sum_{c=1}^C\|\mathbb{E}[h(\mathbf{x}_s)|y_s^c]-\mathbb{E}[h(\mathbf{x}_t)|y_t^c]\|^2$, where $\mathbb{E}[\cdot]$ denotes the expectation of the subspace samples.

More specifically, consider a linear mapping $h(\mathbf{x})=A^{\top}\mathbf{x}$ for the source and the target domains, where $A\in \mathbb{R}^{d \times p}$. (\ref{eq:JDA0}) can then be re-expressed as
\begin{align}
\begin{split}
d(\mathcal{D}_s,\mathcal{D}_t) &\approx \mu_1 \left\|\frac{1}{n_{s}} \sum_{i=1}^{n_{s}} A^{\top} \mathbf{x}_{s, i}-\frac{1}{n_{t}} \sum_{j=1}^{n_{t}} A^{\top} \mathbf{x}_{t, j} \right\|_2^{2} \\
&+ \mu_2 \sum_{c=1}^C {\left \| \frac 1 {n_s^c}\sum_{i=1}^{n_s^c}A^{\top} \mathbf{x}_{s,i}^c - \frac 1 {n_t^c}\sum_{j=1}^{n_t^c}A^{\top}\mathbf{x}_{t,j}^c \right \|_2^2}, \label{eq:MMD}
\end{split}
\end{align}
where $\mathbf{x}_{s,i}^c$ and $\mathbf{x}_{t,j}^c$ are the feature vectors in the $c$-th class of the source domain and the target domain, respectively, and $n_s^c$ and $n_t^c$ are the number of examples in the $c$-th class of the source domain and the target domain, respectively.

When $\mu_1=1$ and $\mu_2=0$, (\ref{eq:MMD}) becomes transfer component analysis (TCA) \cite{pan2011domain}. When $\mu_1=1$ and $\mu_2=1$, (\ref{eq:MMD}) becomes JDA. When $\mu_1=1-\mu_2$, (\ref{eq:MMD}) becomes BDA. Thus, these traditional DA approaches based on the marginal and conditional MMDs with equal or different weights only approximate the joint probability distribution shift.

\subsection{DJP-MMD}

As shown in the previous subsection, the traditional DA approximates the domain discrepancy by a weighted or unweighted sum of the marginal and conditional MMDs. This subsection proposes DJP-MMD, which computes the joint probability discrepancy directly, and maximizes both the domain transferability and the class discriminability.
\begin{dfn}
\textbf{(The Joint Probability Discrepancy)} Let $c=\{1,...,C\}$ and $\hat c=\{1,...,C\}$ be the label sets of the source and the target domains, respectively. Let $P(X|Y)$ be the class-conditional probability, and $P(Y)$ the class prior probability. Then, according to the Bayesian law, the joint probability discrepancy is
\end{dfn}

\begin{align}
d(\mathcal{D}_s,\mathcal{D}_t) &=d\left(P(X_s|Y_s)P(Y_s),P(X_t|Y_t)P(Y_t)\right) \nonumber \\
&=\sum_{c=\hat c}^C \sum_{\hat c=1}^C  d\left(P(X_s|Y_s^c)P(Y_s^c),P(X_t|Y_t^{\hat c})P(Y_t^{\hat c})\right) \nonumber\\
&+\sum_{c\neq \hat c} \sum_{\hat c=1}^C d\left(P(X_s|Y_s^c)P(Y_s^c),P(X_t|Y_t^{\hat c})P(Y_t^{\hat c})\right) \nonumber\\
&=\sum_{c=1}^C  d\left(P(X_s|Y_s^c)P(Y_s^c),P(X_t|Y_t^c)P(Y_t^c)\right) \nonumber\\
&+\sum_{c\neq \hat c} \sum_{\hat c=1}^C d\left(P(X_s|Y_s^c)P(Y_s^c),P(X_t|Y_t^{\hat c})P(Y_t^{\hat c})\right) \nonumber \\
&\equiv \mathcal{M}_T + \mathcal{M}_D \label{eq:JPD}
\end{align}
$\mathcal{M}_T$ (or $\mathcal{M}_D$) measures the joint probability discrepancy on the same class (or between different classes) in the two domains.

The difference between the first line of (\ref{eq:JDA0}) and that of (\ref{eq:JPD}) is that the former is based on the product of the marginal probability and the posterior probability, whereas the latter is based on the product of the class-conditional probability and the class prior probability. Though theoretically they are equivalent, (\ref{eq:JPD}) can be computed directly from the data without approximation, and it enables us to incorporate class discriminability into the discrepancy, as shown later in this subsection.

Directly minimizing (\ref{eq:JPD}) can improve the transferability between the source and the target domains, but it completely ignores the discriminability between different classes, which may not be good for classification. So, we define the \emph{discriminative joint probability discrepancy} as
\begin{align}
d(\mathcal{D}_s,\mathcal{D}_t)= \mathcal{M}_T - \mu \mathcal{M}_D, \label{eq:DJPD}
\end{align}
where $\mu>0$ is a trade-off parameter. $\mathcal{M}_T$ measures the transferability of the same class between different domains, and $\mathcal{M}_D$ measures the discriminability between different classes of different domains.

Next, we introduce specifically how to compute $\mathcal{M}_T$ and $\mathcal{M}_D$ by MMD.

\textbf{MMD for Transferability:} From (\ref{eq:JPD}) we have
\begin{align}
\begin{split}
\mathcal{M}_T &= \sum_{c=1}^C  d\left(P(X_s|Y_s^c)P(Y_s^c),P(X_t|Y_t^c)P(Y_t^c)\right)\\
&=\sum_{c=1}^C \left \|\mathbb{E}[f(\mathbf{x}_s) | y_s^c]P(y_s^c)-\mathbb{E}[f(\mathbf{x}_t) | y_t^c)]P( y_t^c)\right \|^2, \label{eq:MMDt}
\end{split}
\end{align}
where empirically
\begin{align}
\mathbb{E}[f(\mathbf{x}_s) | y_s^c]&=\frac 1 {n_s^c}\sum_{i=1}^{n_s^c}A^{\top}\mathbf{x}_{s,i}^c, \\ P(y_s^c)&=\frac {n_s^c} {n_s}. \label{eq:cp}
\end{align}

Then,
\begin{align}
\mathbb{E}[f(\mathbf{x}_s) | y_s^c]P(y_s^c)=\frac 1 {n_s}\sum_{i=1}^{n_s^c}A^{\top}\mathbf{x}_{s,i}^c. \label{eq:jps}
\end{align}

Similarly, we have
\begin{align}
\mathbb{E}[f(\mathbf{x}_t) | y_t^c]P(y_t^c)=\frac 1 {n_t}\sum_{i=1}^{n_t^c}A^{\top}\mathbf{x}_{t,i}^c,
\label{eq:jpt}
\end{align}
where $y_t$ is target-domain pseudo-label estimated from a classifier trained in the source domain.

Substituting (\ref{eq:jps}) and (\ref{eq:jpt}) into (\ref{eq:MMDt}), we have
\begin{align}
\mathcal{M}_T = \sum_{c=1}^C {\left \| \frac 1 {n_s}\sum_{i=1}^{n_s^c}A^{\top}\mathbf{x}_{s,i}^c - \frac 1 {n_t}\sum_{j=1}^{n_t^c}A^{\top}\mathbf{x}_{t,j}^c \right \|_2^2}. \label{eq:JPMMD}
\end{align}

Note that, the joint probability MMD in (\ref{eq:JPMMD}) is different from the conditional MMD in (\ref{eq:MMD}), since $n_s^c$ and $n_t^c$ are used in (\ref{eq:MMD}), whereas $n_s$ and $n_t$ are used in (\ref{eq:JPMMD}).  $n_t^c$ in (\ref{eq:MMD}) is estimated, whereas $n_t$ in (\ref{eq:JPMMD}) is known precisely and hence more accurate than $n_t^c$.

\textbf{MMD for Discriminability:} From (\ref{eq:JPD}) we have
\begin{align}
\begin{split}
\mathcal{M}_D &= \sum_{c\neq \hat c} \sum_{\hat c=1}^C d(P(X_s|Y_s^c)P(Y_s^c),P(X_t|Y_t^{\hat c})P(Y_t^{\hat c}))\\
&=\sum_{c\neq \hat c} \sum_{\hat c=1}^C \left \|\mathbb{E}[f(\mathbf{x}_s) | y_s^c]P(y_s^c)-\mathbb{E}[f(\mathbf{x}_t) | y_t^{\hat c}]P(y_t^{\hat c})\right \|^2.
\end{split}
\end{align}

Using the same derivation as before, it follows that
\begin{align}
\mathcal{M}_D = \sum_{c\neq \hat c} \sum_{\hat c=1}^C {\left \| \frac 1 {n_s}\sum_{i=1}^{n_s^c}A^{\top}\mathbf{x}_{s,i}^c - \frac 1 {n_t}\sum_{j=1}^{n_t^{\hat c}}A^{\top}\mathbf{x}_{t,j}^{\hat c} \right \|_2^2}. \label{eq:JPMMD2}
\end{align}

\textbf{The DJP-MMD:} Let the source domain one-hot coding label matrix be $Y_s=[\mathbf{y}_{s,1}; \cdots; \mathbf{y}_{s,n_s}]$, and the predicted target domain one-hot coding label matrix be $\hat{Y}_t=[\hat{\mathbf{y}}_{t,1}; \cdots; \hat{\mathbf{y}}_{t,n_t}]$, where $\mathbf{y}_{s,i}\in \mathbb{R}^{1 \times C}$ and $\hat{\mathbf{y}}_{t,i}\in \mathbb{R}^{1 \times C}$. Then, (\ref{eq:JPMMD}) can be re-expressed as
\begin{align}
\mathcal{M}_T = \left \| A^{\top}X_sN_s-A^{\top}X_tN_t \right \|_F^2,
\end{align}
where $N_s$ and $N_t$ are defined as
\begin{align}
N_s=\frac {Y_s} {n_s},\ N_t=\frac {\hat Y_t} {n_t}. \label{eq:Nst}
\end{align}
The $c$-th column of $A^{\top}X_sN_s\in\mathbb{R}^{p\times C}$ (or $A^{\top}X_tN_t$) is the mean mapped feature of Class~$c$ in the source (or target) domain.

Define
\begin{align}
F_s&=[Y_s(:,1)*(C-1), ..., Y_s(:,C)*(C-1)], \nonumber \\
\hat F_t&=[\hat{Y}_t(:,1:C)_{\hat c\neq 1}, ..., \hat{Y}_t(:,1:C)_{\hat c\neq C}],
\end{align}
where $Y_s(:,c)$ denotes the $c$-th column of $Y_s$, $Y_s(:,c)*(C-1)$ repeats $Y_s(:,c)$ $C-1$ times to form a matrix in $\mathbb{R}^{n_s\times (C-1)}$, and $\hat{Y}_t(:,1:C)_{\hat c\neq 1}$ is formed by the 1st to the $C$-th (except the 1st) columns of $\hat{Y}_t$. Clearly, $F_s\in \mathbb{R}^{n_s\times (C(C-1))}$ and $\hat F_t\in \mathbb{R}^{n_t\times (C(C-1))}$. $F_s$ is fixed, and $\hat{F}_t$ is constructed from the pseudo labels, which are updated iteratively.

Then, (\ref{eq:JPMMD2}) can be re-expressed as
\begin{align}
\mathcal{M}_D = \left \| A^{\top}X_sM_s-A^{\top}X_tM_t \right \|_F^2,
\end{align}
where
\begin{align}
M_s=\frac {F_s} {n_s},\ M_t=\frac {\hat F_t} {n_t}. \label{eq:Mst}
\end{align}

To facilitate DA, we need to minimize $d(\mathcal{D}_s,\mathcal{D}_t)$ in (\ref{eq:DJPD}), i.e., we solve the optimal linear mapping $A$ by
\begin{align}
\begin{split}
\min_{A} \ & \left \| A^{\top}X_sN_s-A^{\top}X_tN_t \right \|_F^2 \\
&-\mu \left \| A^{\top}X_sM_s-A^{\top}X_tM_t \right \|_F^2
\end{split} \label{eq:DJP-MMD}
\end{align}

DJP-MMD in (\ref{eq:DJP-MMD}) has two appealing properties: 1) it considers the joint probability MMD directly, which in theory is more accurate than considering the marginal MMD and conditional MMD separately; and, 2) it improves the domain transferability and the class discriminability simultaneously.

\subsection{Use DJP-MMD in DA}

To verify the superiority of the proposed DJP-MMD over the traditional MMDs, we embed it into an unsupervised joint probability DA (JPDA) framework with a regularization term and a principal component preservation constraint, which have also been used in the classical TCA and JDA. More specifically,
\begin{align}
\begin{split}
\min_{A}& \ \left \| A^{\top}X_sN_s-A^{\top}X_tN_t \right \|_F^2\\
& - \mu \left \| A^{\top}X_sM_s-A^{\top}X_tM_t \right \|_F^2+ \lambda \|A\|_F^2\\
&s.t.\ A^{\top}X H X^{\top}A=I, \label{eq:loss}
\end{split}
\end{align}
where $H=I-\mathbf{1}_n $ is the centering matrix, in which $n=n_s+n_t$ and $\mathbf{1}_n\in \mathbb{R}^{n\times n}$ is a matrix with all elements being $\frac 1 n$.

\subsection{Optimize the JPDA}

Define $X=[X_s, X_t]$. We can write the Lagrange function \cite{bishop2006pattern} of (\ref{eq:loss}) as
\begin{align}
\begin{split}
\mathcal{J} &=\operatorname{tr} \left( A^{\top} \left(X(R_{\min}-\mu R_{\max})X^{\top}+\lambda I\right) A \right) \\
& + \operatorname{tr} \left(\eta(I-A^{\top}X H X^{\top} A)\right), \label{eq:J}
\end{split}
\end{align}
where
\begin{align}
R_{\min}&=\begin{bmatrix} N_sN_s^{\top} & -N_sN_t^{\top} \\ -N_tN_s^{\top} & N_tN_t^{\top} \end{bmatrix},  \label{eq:Rmin}\\
R_{\max}&=\begin{bmatrix} M_sM_s^{\top} & -M_sM_t^{\top} \\ -M_tM_s^{\top} & M_tM_t^{\top} \end{bmatrix}. \label{eq:Rmax}
\end{align}
$R_{\max}$ has dimensionality $n\times n$, which does not change with the number of classes.

By setting the derivative $\nabla_A \mathcal{J}=\mathbf{0}$, (\ref{eq:J}) becomes a generalized eigen-decomposition problem:
\begin{align}
\left(X(R_{\min}-\mu R_{\max})X^{\top}+\lambda I\right)A = \eta X H X^{\top} A.  \label{eq:eigen}
\end{align}
$A$ is then formed by the $p$ trailing eigen-vectors. A classifier can then be trained on $A^{\top}X_s$ and applied to $A^{\top}X_t$.

The pseudocode of JPDA for classification is summarised in Algorithm~\ref{alg:1}.

\begin{algorithm}  \caption{Joint Probability Distribution Adaptation (JPDA)}
\label{alg:1}
\KwIn {$X_s$ and $X_t$, source and target domain feature matrices;\\
\hspace*{10mm} $Y_s$, source domain one-hot coding label matrix;\\
\hspace*{10mm} $p$, subspace dimensionality;\\
\hspace*{10mm} $\mu$, trade-off parameter;\\
\hspace*{10mm} $\lambda$, regularization parameter;\\
\hspace*{10mm} $T$, number of iterations.}
\KwOut {$\hat Y_t$, estimated target domain labels.}
\For{$n=1,...,T$}
{
    Construct the joint probability matrix $R_{\min}$ and $R_{\max}$ by (\ref{eq:Rmin}) and (\ref{eq:Rmax})\;
    Solve the generalized eigen-decomposition problem in (\ref{eq:eigen}) and select the $p$ trailing eigenvectors to construct the projection matrix $A$\;
    Train a classifier $f$ on $(A^{\top}X_s,Y_s)$\ and apply it to $A^{\top}X_t$ to obtain $\hat{Y}_t$.
}
\end{algorithm}

\subsection{Kernelization}

To consider nonlinear DA, kernel function $\phi: \mathbf{x} \mapsto \phi(\mathbf{x})$ in an RKHS can be adopted. We then have $K_s=\Phi(X)^{\top}\Phi(X_s)$, $K_t=\Phi(X)^{\top}\Phi(X_t)$, and $K=[K_s, K_t]$, where $\Phi(X)=[\phi(\mathbf{x}_1),..,\phi(\mathbf{x}_n)]$, and $n=n_s+n_t$.

Then, the objective function becomes
\begin{align}
\begin{split}
\min_{A}& \ \left \| A^{\top}K_sN_s-A^{\top}K_tN_t \right \|_F^2\\
& - \mu \left \| A^{\top}K_sM_s-A^{\top}K_tM_t \right \|_F^2+ \lambda \|A\|_F^2 \\
&s.t.\ A^{\top}K H K^{\top}A=I, \label{eq:kloss}
\end{split}
\end{align}
(\ref{eq:kloss}) can be optimized in a similar way to (\ref{eq:J}).

\subsection{Computational Complexity}

The most computationally expensive operations in Algorithm~1 are generalized eigen-decomposition and the MMD matrices construction.

For most practical applications, both $T$ (the number of iterations) and $p$ (the subspace dimensionality) are much smaller than $\min(d,n)$. The computational cost of solving the generalized eigen-decomposition problem for dense matrices is $\mathbf{O}(Tpd^2)$, of constructing the MMD matrices is $\mathbf{O}(Tn^2)$, and of all other steps is $\mathbf{O}(Tdn)$. Thus, the total theoretical computational complexity is $\mathbf{O}(Tpd^2+Tn^2+Tdn)$. The empirical computational complexity will be given in Section~IV.

\section{Experiments}

Experiments are performed in this section to demonstrate the performance of JPDA. The code is available at https://github.com/chamwen/JPDA.

\subsection{Datasets}

\begin{figure}[htpb]  \centering
\includegraphics[width=0.45\textwidth,clip]{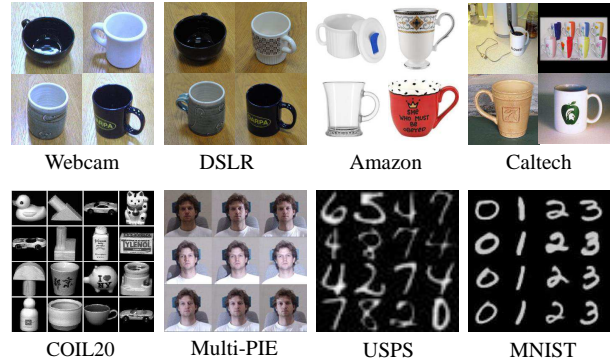}
\caption{Sample images from the six datasets. Webcam, DSLR and Amazon are all from the Office dataset.}
\label{fig:dataset}
\end{figure}

Office, Caltech, COIL, Multi-PIE, MNIST and USPS are six benchmark datasets widely used to evaluate visual DA algorithms. They were also used in our experiments. Some examples from these datasets are shown in Fig.~\ref{fig:dataset}.

\textbf{Object Recognition}: Office+Caltech \cite{griffin2007caltech} is a popular benchmark for visual DA. It contains four real-world object domains: Caltech ($\textit{C}$), Amazon ($\textit{A}$), Webcam ($\textit{W}$), and DSLR ($\textit{D}$). Our experiments used the public Office+Caltech dataset with SURF features released in \cite{gong2012geodesic}. By randomly selecting one domain as the source domain and a different domain as the target domain, we had $4\times 3=12$ different cross-domain transfer tasks.

COIL contains 20 objects with 1,440 images. The images of each object were taken 5 degrees apart as the object was rotated on a turntable, and each object has 72 images of $32\times 32$ pixels. The dataset was partitioned into two equal subsets (COIL1 and COIL2) with different distributions.

\textbf{Face Recognition}: Multi-PIE is a benchmark for face recognition. The database has 68 individuals with 41,368 $32\times 32$ face images. It has five subsets: C05 (left pose), C07 (upward pose), C09 (downward pose), C27 (frontal pose), and C29 (right pose). In each subset (pose), all face images were taken under different lighting, illumination, and expression conditions. By randomly selecting one subset (pose) as the source domain and a different one as the target domain, we had $5 \times 4=20$ different cross-domain transfer tasks.

\textbf{Digit Recognition}: USPS and MNIST are two public digit recognition datasets with different resolutions. Our experiments used the public USPS and MNIST datasets released by Long \emph{et al.} \cite{long2013transfer}, which randomly sampled 1,800 images in USPS and 2,000 images in MNIST. They both have 10 classes of digits, with different distributions.

\subsection{Algorithms}

To validate the effectiveness of the proposed DJP-MMD, we compared JPDA with three unsupervised DA approaches, TCA \cite{pan2011domain}, JDA \cite{long2013transfer} and BDA (which used the $\mathcal{A}$-distance \cite{wang2018visual} to compute the weight, instead of grid search in \cite{wang2017balanced}). Because they have different MMD metrics but the same regularization term, we can attribute the performance differences solely to the MMD metrics.

A 1-nearest neighbor classifier was applied after TCA, JDA, BDA and JPDA. The parameter settings in \cite{long2013transfer} were used for TCA, JDA and BDA. We fixed $p=100$ and $T=10$ in all experiments, and the regularization parameter $\lambda=1$ with linear kernel for Office+Caltech dataset, $\lambda=0.1$ with primal kernel for other datasets. $\mu=0.1$ was used in JPDA.

\subsection{Results}

The target domain classification accuracy was used as the performance measure.

The classification accuracies of the four algorithms are given in Table~\ref{tab:results}. JPDA outperformed the three baselines in most tasks, and its average performance was also the best, suggesting that JPDA can obtain a more transferrable and also more discriminative feature mapping for cross-domain visual adaptation. Although the $\mathcal{A}$-distance based BDA was proposed to improve JDA by adding a balance factor between the marginal MMD and the conditional MMD, it did not demonstrate better performance in our experiments.
\begin{table}[htpb]  \centering  \setlength{\tabcolsep}{2mm}
  \caption{Classification accuracy (\%) of the four algorithms.}
    \begin{tabular}{c|cc|cccc}  \toprule
    Dataset & Source & Target & TCA   & JDA   & BDA   & JPDA \\  \midrule
    \multirow{20}{*}{Multi-PIE}
    & \multirow{4}[2]{*}{C05}
           & C07   & 40.76 & 58.81 & 58.20 & \textbf{59.36} \\
      &    & C09   & 41.79 & 54.23 & 52.82 & \textbf{66.67} \\
      &    & C27   & 59.63 & \textbf{84.50} & 83.03 & 83.99 \\
      &    & C29   & 29.35 & \textbf{49.75} & 49.14 & 49.51 \\ \cline{2-7}
    & \multirow{4}[2]{*}{C07}
           & C05   & 41.81 & 57.62 & 57.35 & \textbf{63.00} \\
      &    & C09   & 51.47 & \textbf{62.93} & 62.75 & 60.85 \\
      &    & C27   & 64.73 & 75.82 & 75.76 & \textbf{77.05} \\
      &    & C29   & 33.70 & 39.89 & 39.71 & \textbf{47.67} \\ \cline{2-7}
    & \multirow{4}[2]{*}{C09}
           & C05   & 34.69 & 50.96 & 51.35 & \textbf{59.78} \\
      &    & C07   & 47.70 & 57.95 & 56.41 & \textbf{63.35} \\
      &    & C27   & 56.23 & 68.46 & 67.86 & \textbf{74.47} \\
      &    & C29   & 33.15 & 39.95 & 42.40 & \textbf{52.70} \\ \cline{2-7}
    &  \multirow{4}[2]{*}{C27}
           & C05   & 55.64 & 80.58 & 80.52 & \textbf{84.87} \\
      &    & C07   & 67.83 & 82.63 & 83.06 & \textbf{83.24} \\
      &    & C09   & 75.86 & 87.25 & 87.25 & \textbf{87.44} \\
      &    & C29   & 40.26 & 54.66 & 54.53 & \textbf{65.38} \\ \cline{2-7}
    & \multirow{4}[2]{*}{C29}
           & C05   & 26.98 & 46.46 & 47.99 & \textbf{53.63} \\
      &    & C07   & 29.90 & 42.05 & 43.22 & \textbf{51.32} \\
      &    & C09   & 29.90 & 53.31 & 47.92 & \textbf{55.76} \\
      &    & C27   & 33.64 & 57.01 & 57.10 & \textbf{58.49} \\ \midrule
    \multirow{12}{*}{Office+Caltech}
    & \multirow{3}[2]{*}{C}
           & A  & 38.20 & 44.78 & 44.57 & \textbf{47.60} \\
      &    & W     & 38.64 & 41.69 & 40.34 & \textbf{45.76} \\
      &    & D     & 41.40 & 45.22 & 45.22 & \textbf{46.50} \\ \cline{2-7}
    & \multirow{3}[2]{*}{A}
           & C     & 37.76 & 39.36 & 39.27 & \textbf{40.78} \\
      &    & W     & 37.63 & 37.97 & 37.97 & \textbf{40.68} \\
      &    & D     & 33.12 & 39.49 & \textbf{40.76} & 36.94 \\ \cline{2-7}
    & \multirow{3}[2]{*}{W}
           & C     & 29.30 & 31.17 & 31.43 & \textbf{34.55} \\
      &    & A     & 30.06 & 32.78 & 32.46 & \textbf{33.82} \\
      &    & D     & 87.26 & \textbf{89.17} & \textbf{89.17} & 88.54 \\ \cline{2-7}
    & \multirow{3}[2]{*}{D}
           & C     & 31.70 & 31.52 & 31.17 & \textbf{34.73} \\
      &    & A     & 32.15 & 33.09 & 33.19 & \textbf{34.66} \\
      &    & W     & 86.10 & 89.49 & 89.49 & \textbf{91.19} \\ \midrule
    \multirow{2}{*}{COIL}
      & COIL1 & COIL2  & 88.47 & 89.31 & 89.44 & \textbf{92.08} \\
      & COIL2 & COIL1  & 85.83 & 88.47 & 88.33 & \textbf{89.86} \\  \midrule
    \multirow{2}{*}{USPS+MNIST}
      & USPS  & MNIST  & 51.05 & 59.65 & \textbf{59.90} & 59.20 \\
      & MNIST & USPS  & 56.28 & 67.28 & 67.39 & \textbf{68.94} \\ \midrule
    \multicolumn{3}{c|}{Average} & 47.22 & 57.37 & 57.18 & \textbf{60.68} \\ \bottomrule
    \end{tabular}   \label{tab:results}
\end{table}

\begin{figure}  \centering
\includegraphics[width=\linewidth,trim=50 0 45 0,clip]{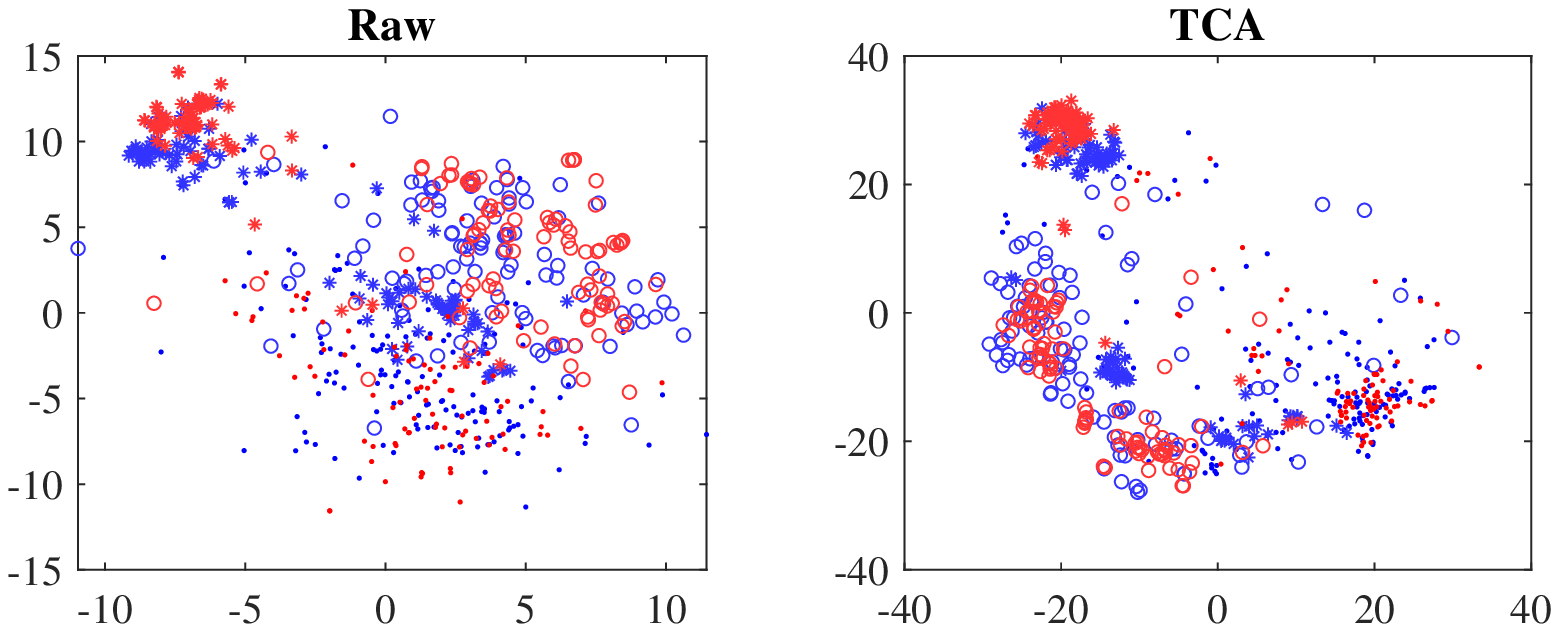}
\includegraphics[width=\linewidth,trim=50 0 45 0,clip]{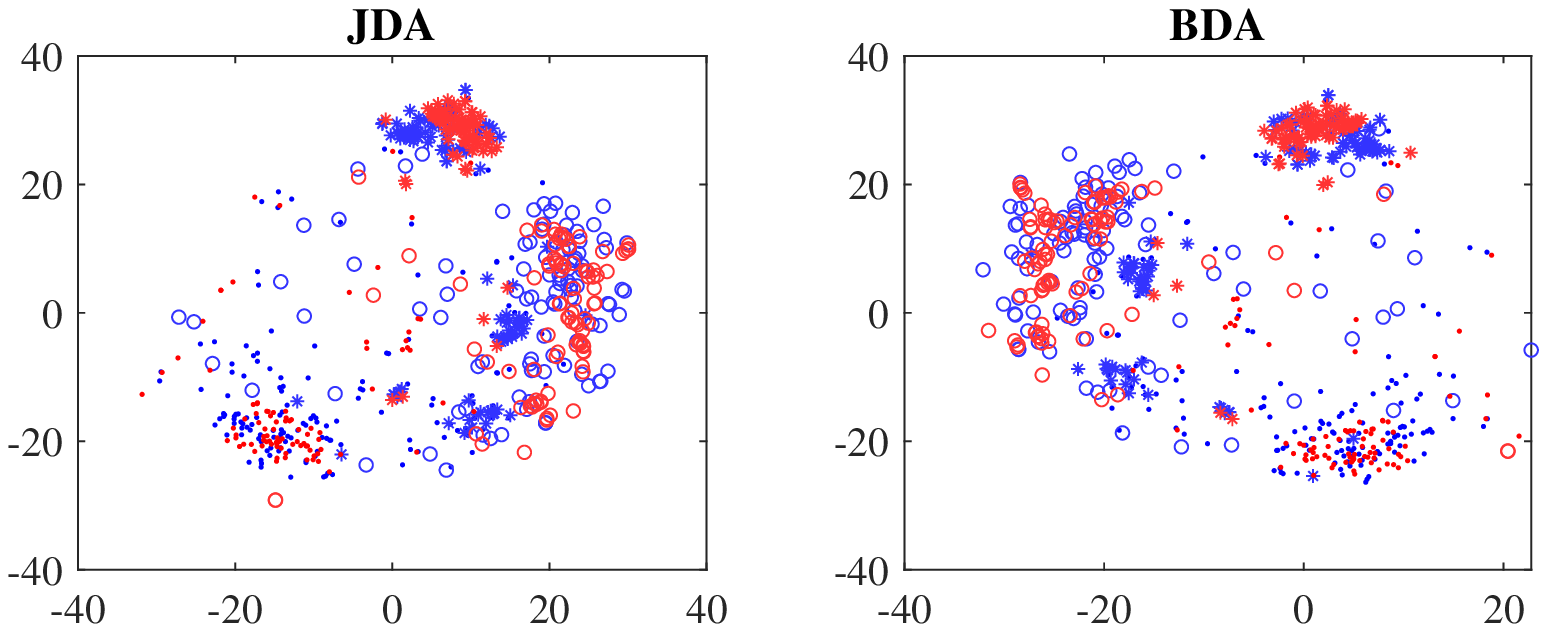}
\includegraphics[width=\linewidth,trim=50 0 45 0,clip]{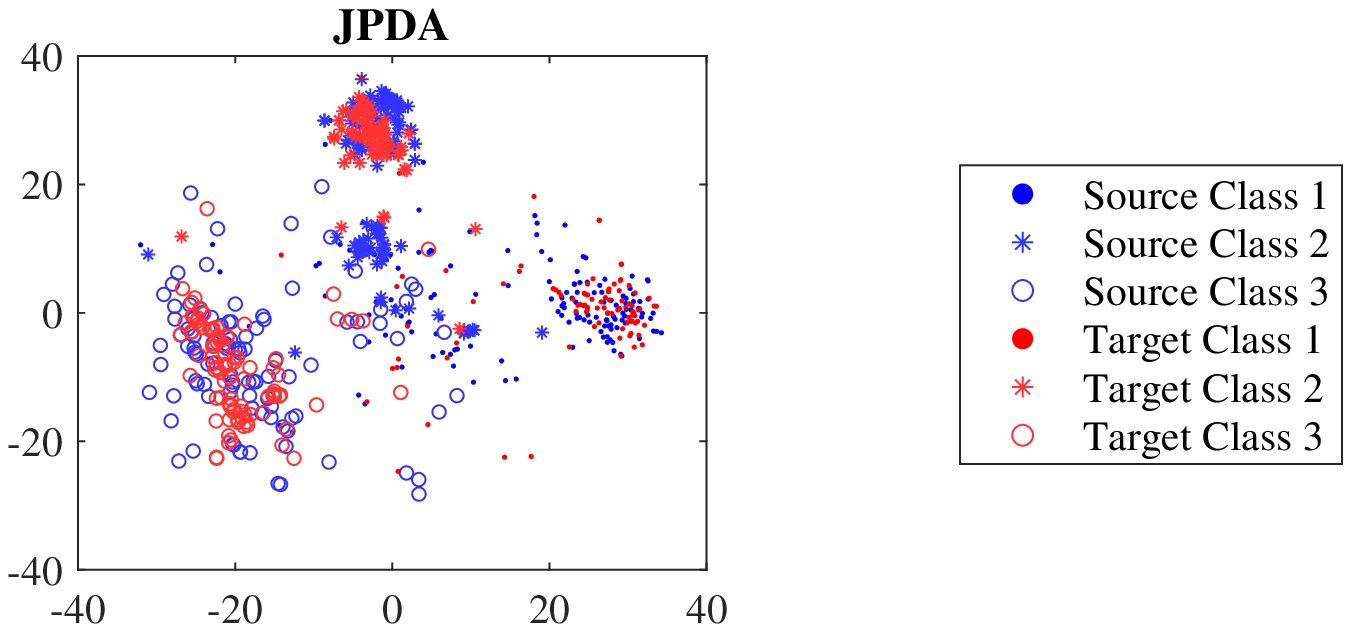}
\caption{$t$-SNE visualization of the first three classes' data distributions before and after different DA approaches, when transferring Caltech (source) to Amazon (target).}
\label{fig:view}
\end{figure}

We also verified whether JPDA can increase both the transferability and the discriminability. We used $t$-SNE \cite{maaten2008} to reduce the dimensionality of the feature to two, and visualize the data distributions. Fig.~\ref{fig:view} shows the results of the first three classes' data distributions when transferring Caltech (source) to Amazon (target), before and after different distribution adaptation approaches, where \emph{Raw} denotes the raw data distribution. For the raw distribution, the samples from Class~1 and Class~3 (also some from Class~2) of the source and the target domains are mixed together. After DA, JPDA brings data distributions of the source and the target domains together, and also keeps samples from different classes well-separated. JDA and BDA do not have such good discriminability, especially for samples from Classes~2 and 3.

\subsection{Convergence and Time Complexity}

\begin{figure}[htpb]  \centering
  \subfigure[]{\label{view.mmd}
    \includegraphics[width=0.34\textwidth,trim=3 0 17 10,clip]{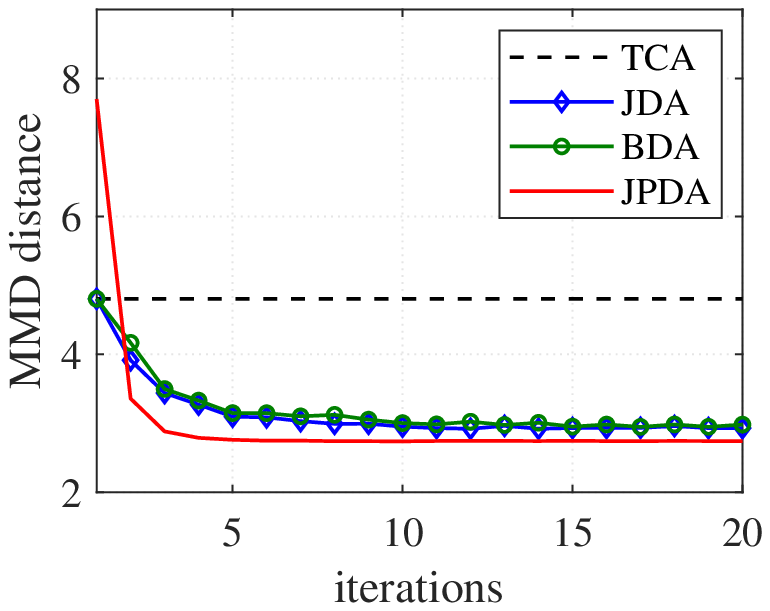} }
  \subfigure[]{\label{view.acc}
    \includegraphics[width=0.34\textwidth,trim=3 0 17 9,clip]{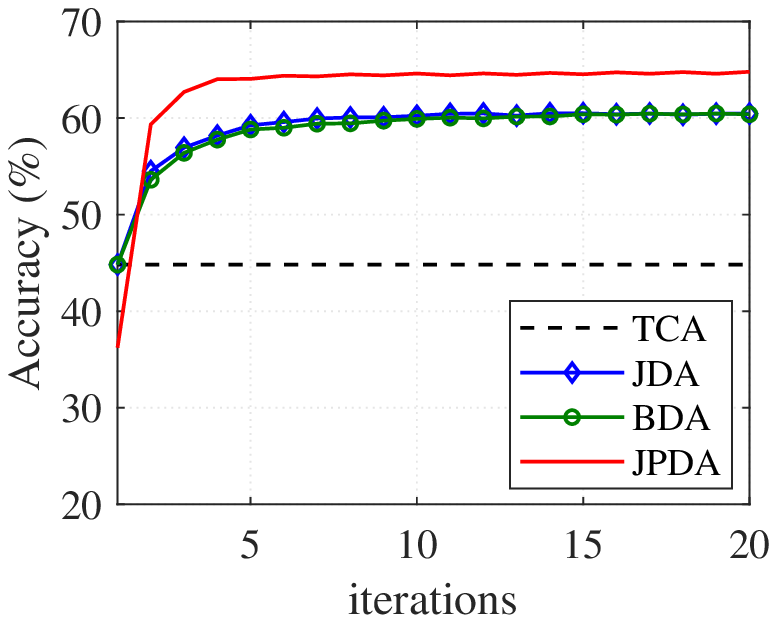} }
\caption{(a) Average MMD distances and (b) average classification accuracies of different DA approaches w.r.t. the number of training iterations, in the 20 Multi-PIE tasks.} \label{fig:conv}
\end{figure}

We then empirically checked the convergence of different DA approaches. Fig.~\ref{fig:conv} shows the average MMD distances (the method to compute the distance can be found in \cite{long2013transfer}) and classification accuracies in the 20 transfer tasks on Multi-PIE, as the number of iterations increased from 1 to 20. JPDA converged quickly and achieved a much smaller MMD distance, as well as a higher accuracy.

The computational costs of the four algorithms are shown in Table~\ref{tab:time}. JPDA was always faster than JDA and BDA. Especially, when the dataset is large (Multi-PIE), JPDA can save over 50\% computing time. TCA was the fastest, since it is not iterative. BDA was the most time-consuming approach, because it needed to train $C+1$ classifiers to compute the balance factor.

\begin{table}[htpb]  \centering
  \caption{Computational cost (seconds) of different approaches.}
    \begin{tabular}{c|cccc}  \toprule
     & TCA   & JDA   & BDA   & JPDA \\  \midrule
    C05$\to$C07 & \textbf{2.58} & 94.46 & 107.47 & \underline{46.12} \\
    C$\to$A & \textbf{2.93} & 31.61 & 34.73 & \underline{30.65} \\
    MNIST$\to$USPS  & \textbf{0.75} & 9.04  & 13.58 & \underline{8.41} \\  \bottomrule
    \end{tabular}   \label{tab:time}
\end{table}

\subsection{Parameters Sensitivity}

\begin{figure}[htpb]  \centering
  \subfigure[]{\label{view.para_mu}
    \includegraphics[width=0.35\textwidth,trim=10 2 33 15,clip]{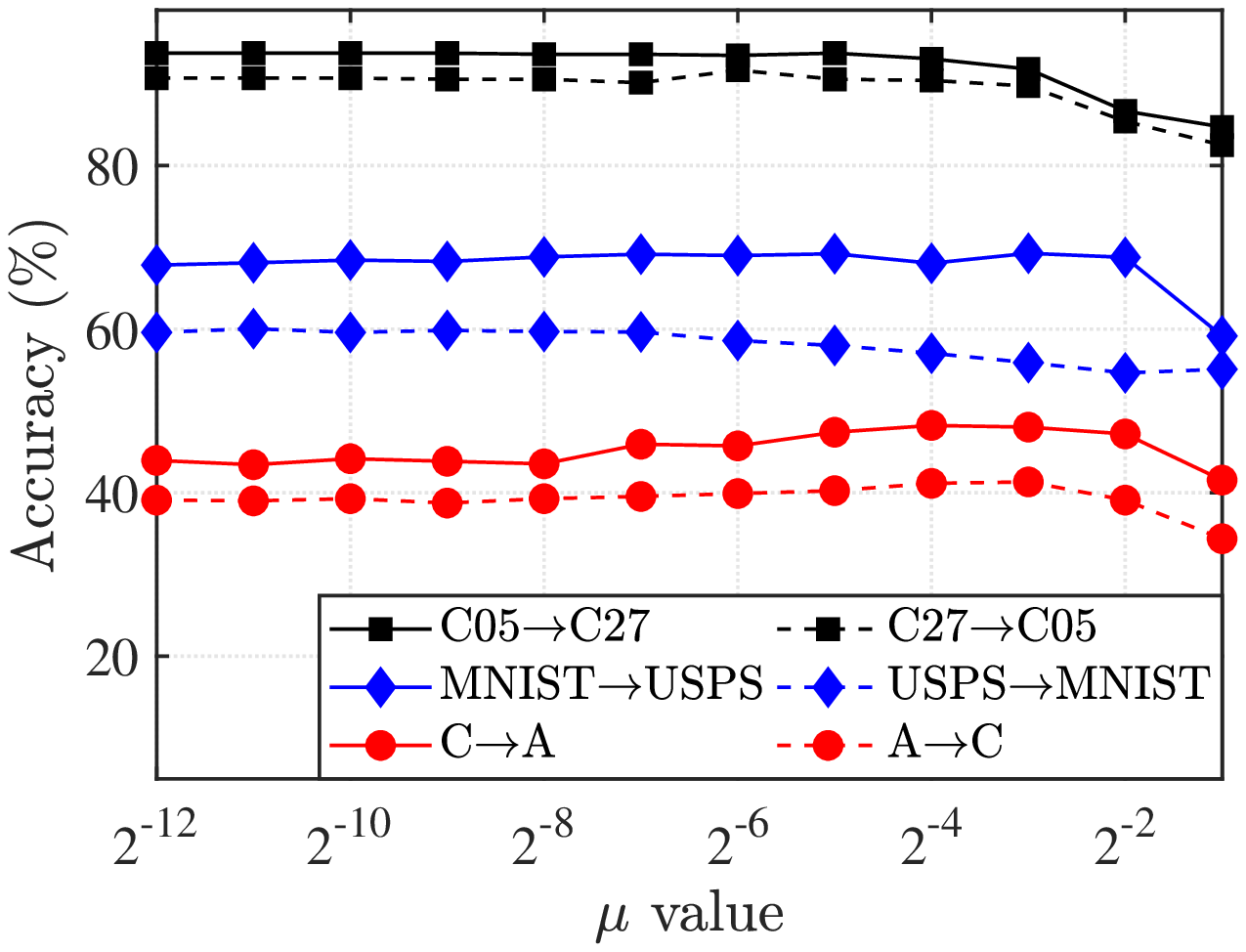} }
  \subfigure[]{\label{view.para_lambda}
    \includegraphics[width=0.35\textwidth,trim=10 2 33 10,clip]{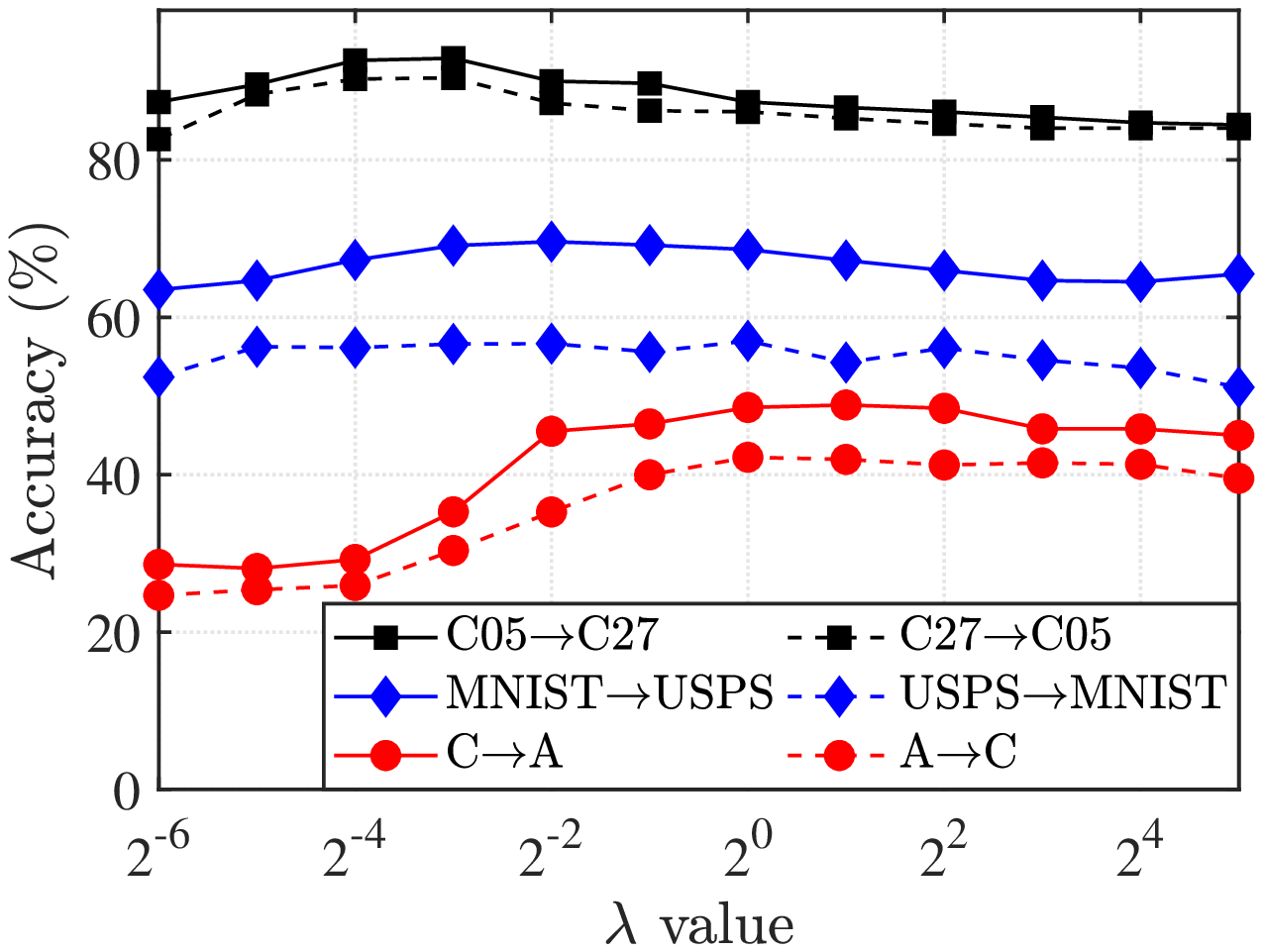} }
\caption{Average classification accuracies of JPDA in six tasks w.r.t. (a) the trade-off parameter $\mu$, and, (b) the regularization parameter $\lambda$.} \label{fig:para}
\end{figure}

We also analyzed the parameter sensitivity of JPDA on different datasets to validate that a wide range of parameter values can be used to obtain satisfactory performance. Two main adjustable parameters, the trade-off parameter $\mu$ and the regularization parameter $\lambda$, were studied. The results are shown in Fig.~\ref{fig:para}. JPDA is robust to $\mu$ in $[0.001, 0.2]$ and $\lambda$ in $[0.01,10]$.

\subsection{Ablation Study}

\begin{figure}[htpb]  \centering
\includegraphics[width=0.49\textwidth,trim=5 0 5 10,clip]{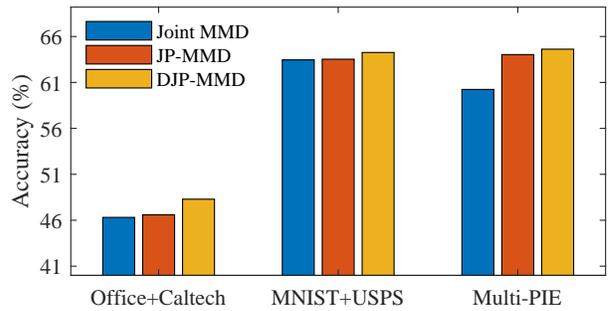}
\caption{Average classification accuracies when different MMDs are used in DA.} \label{fig:ablation}
\end{figure}

Next, we conducted ablation study to check if the discriminative MMD $\mathcal{M}_D$ can indeed improve the discriminability in the target domain, i.e., with $\mathcal{M}_D$ (DJP-MMD) and without $\mathcal{M}_D$ (JP-MMD, which only considers the transferability). The joint MMD was also used as a baseline. When embedded in DA, the average classification accuracies of the three MMDs are shown in Fig.~\ref{fig:ablation}. On average, JP-MMD outperformed the joint MMD, and DJP-MMD, which further considers the discriminability, achieved the best classification performance.

\section{Conclusion}

This paper has proposed simple yet effective DJP-MMD for DA. We verified its performance by embedding it into a JPDA framework. JPDA improves the transferability between different domains and the discriminability between different classes simultaneously, by minimizing the joint probability MMD of the same class in the source and target domains (i.e., increase the domain transferability), and maximizing the joint probability MMD of different classes (i.e., increase the class discriminability). Compared with the traditional MMD based approaches, JPDA is simpler, and more effective in measuring the discrepancy between different domains. Experiments on six image classification datasets verified the superiority of JPDA.

Our future research will extend DJP-MMD to deep learning and adversarial learning.


\end{document}